\documentclass[10pt,twocolumn,letterpaper]{article}

\usepackage[pagenumbers]{cvpr} 

\usepackage{cuted} 

\usepackage{currfile} 

\usepackage{caption} 
\usepackage{xurl}  
\usepackage{array}  
\usepackage{tabularx}  
\usepackage{colortbl}  

\newcommand{\newc}[1]{{\color{yellow}#1}}  
\renewcommand{\newc}[1]{#1}  
\newcommand{\chg}[1]{{\color{blue}#1}}  
\renewcommand{\chg}[1]{#1}  
\newcommand{\chgp}[1]{{\color{chgpurple}#1}}  
\renewcommand{\chgp}[1]{#1}  

\usepackage{microtype}
\usepackage{amsmath,amssymb,amsthm}
\usepackage{array}  
\usepackage{enumitem}  
\theoremstyle{definition}

\makeatletter
\renewcommand\paragraph{\@startsection{paragraph}{4}{\z@}%
  {1.0ex \@plus0.3ex \@minus0.2ex}{-1em}{\normalfont\normalsize\bfseries}}
\makeatother

\definecolor{lossred}{HTML}{D55E00}
\definecolor{winblue}{HTML}{2B6CB0}
\definecolor{chgpurple}{HTML}{7E22CE}  

\definecolor{cvprblue}{rgb}{0.21,0.49,0.74}
\usepackage[pagebackref,breaklinks,hidelinks]{hyperref}

\def\confName{CVPR}
\def\confYear{2027}

\title{Text Alone Predicts Which Classes\\a Segmentation Readout Keeps}

\author{David Yan\\
Southeast University\\
{\tt\small davidyan@seu.edu.cn}
}

\begin{document}
\maketitle

\begin{abstract}
An open-vocabulary segmentation
readout silently loses a class deployed
beyond its training categories. It still emits \chg{masks with high
confidence}, but the class never wins a single pixel. The training loss falls
anyway, and mIoU can even rise while classes die.
We show that which classes \emph{collapse} is written in the class-name list
alone. \chg{Before any image is seen, we rank each class's collapse risk from
its text similarity to the training vocabulary alone.} The ranking holds at AUROC $0.69$--$0.96$ across nine
model families, twelve backbones and eight pretraining recipes.
\chg{Two further families, measured only after the predictions were written
down, land in the same band.}
What the list writes, it can rewrite.
\chg{With identical images and} \chgp{a class's counterpart moved in or out of
the training list,}\chg{ retraining flips that class's gain over the frozen baseline by $21$
IoU points; swapping in unrelated names changes nothing.} We find a
\emph{vocabulary-distance mechanism} behind the collapse. \chg{Collapsed classes
lose every vote to near-name neighbors, but in the frozen features they
survive.} Turning
down the readout's \newc{amplitude}, or stopping training early, brings them
back, up to $+16$ points, without retraining; trained with its \newc{amplitude}
held down, the readout never collapses.
\end{abstract}

\begin{figure}[t]
\centering
\includegraphics[width=0.90\linewidth]{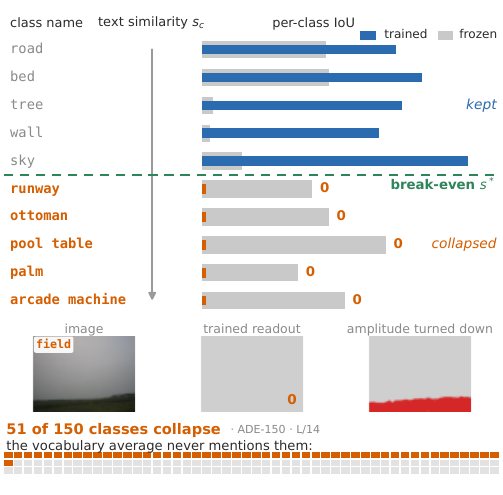}
\caption{\textbf{The list alone says which classes die.} \chgp{ADE-150
classes sorted by text similarity to the training list; below the
break-even point they win no pixels (red), and turning the readout's
amplitude down brings them back.}}
\label{fig:teaser}
\end{figure}
\section{Introduction}
\label{sec:intro}

\chgp{Open-vocabulary segmentation promises to segment every class on the
deployer's list}: a hospital's own taxonomy, a warehouse's inventory labels, and
classes no benchmark covers. \chgp{Each name on such a list is someone's
request.}
The masks come back with high confidence, \chg{but} some of the
classes never appear \chg{in the output}. \chgp{No standard metric reports it.} We call this
\emph{vocabulary-support collapse}. On ADE-150,
51 of the 150 classes collapse (Fig.~\ref{fig:teaser}). \chgp{Ask for all
$150$, and a third never appear.} \chg{No training list keeps every
class alive, since every list leaves some classes out and every deployment
brings new ones. \chgp{Before} deployment, no one can tell which classes will
die. \chgp{Finding out means deploying the model, then checking it on new images
with new labels. The dead classes surface only once the model ships.}} The hospital
never sees the class it asked for\chgp{, and nothing tells it}.

The training vocabulary is \chgp{a second list}, written before training begins. It
already says which classes \chgp{will} die. The readout, a trained mask head over frozen
features, scores a pixel as $\langle F_u,t_c\rangle$. \chgp{Frozen features can already be read with arbitrary class names},
though at coarse spatial \chg{resolution}~\cite{zhou2022maskclip,wang2024sclip}.
\chgp{Training raises the seen classes' scores}~\cite{xu2023san,cho2024catseg}\chgp{, and every}
deployment class \chgp{is either} helped or hurt. \chgp{How close a class's name
sits to the nearest training name decides whether the readout keeps it.} Is a trained readout's ``help-or-hurt'' predictable from
text alone, and controllable through the training vocabulary? \chgp{Both become possible without deployment, new images, or new
labels, because a vocabulary enters the readout only through its text embeddings.} A single
\emph{text similarity}, $s_c=\max_j\cos(t_c,t_j^{\mathrm{train}})$, \chgp{the proximity
to the nearest training name, ranks
each class's collapse risk} before any image is seen.
\chgp{The ranking even has a break-even point
that the list itself predicts. Below the point,} \chg{the readout loses the class and
keeps its confidence}. \chgp{The list also causes the deaths it predicts: with the images held fixed,
moving one name in or out flips that class between kept and collapsed.} \chgp{The two lists already answer both questions.}

Our contributions.
\begin{itemize}[nosep,leftmargin=1.2em]
  \item \textbf{Text alone predicts which classes a readout keeps.}
  We show that a class's text similarity to the training vocabulary, the
  first text-only predictor of per-class collapse, ranks its collapse risk
  across nine families, twelve backbones, and eight pretraining recipes.
  \chg{The predictions were pre-registered before measurement.}
  \item \textbf{The training vocabulary \emph{causes} the collapse.}
  \chg{We run the first experiment that treats a training vocabulary as the
  intervention axis: with the images held fixed, \chgp{moving a class's
  counterpart in or out of the training list} flips that class by $21$ IoU points; swapping in unrelated names
  changes nothing.} The training vocabulary becomes a \chg{design} choice.
  \item \textbf{Collapsed classes are outvoted, and the vote has a knob.}
  We trace the collapse to the vote itself: pixels go to the
  highest-scoring name; training turns
  the readout up along \chgp{the seen names}, until far classes lose every
  vote to near-name neighbors; in the frozen features they survive. One appended name steals the same vote and drains
  a victim class.
  The knob is the
  readout's \newc{amplitude}: turning it down brings the lost classes back, up to
  $+16$ points, without retraining; held down during training, it prevents
  the collapse.
\end{itemize}

\noindent \chgp{One mechanism runs through all three results: what the list
predicts, what one moved name rewrites, and what the knob returns.} We call
it the \emph{vocabulary-distance mechanism} (VDM). \chg{The training
vocabulary itself} is the object of
study, much as datamodels made the individual training sample
one~\cite{ilyas2022datamodels}: here it is measured, attacked, and
intervened on \emph{with images held fixed}. \chgp{A} readout
deployed past its training vocabulary can be
audited \emph{before} the first image runs. \chgp{The two text files alone}
map which classes will die, and a labeled slice certifies the
rest. Every analysis built on a dead class inherits its absence. We release
the audit itself as
VDM-Bench.\footnote{Benchmark and audit cards are attached as supplementary
material.}
Which classes die has always been written in the training vocabulary; now
it can be read and rewritten before anything runs.

\section{Related Work}
\label{sec:related}

\paragraph{Two families, one unmeasured axis.}
\looseness=-2
OVSS readouts are either training-free surgeries on frozen
CLIP~\cite{radford2021clip} (MaskCLIP~\cite{zhou2022maskclip},
SCLIP~\cite{wang2024sclip}, NACLIP~\cite{hajimiri2025naclip},
ProxyCLIP~\cite{lan2024proxyclip}, DouC~\cite{douc2026}) or trained heads
adding adapters and cost aggregation (SAN~\cite{xu2023san},
CAT-Seg~\cite{cho2024catseg}, SED~\cite{xie2024sed},
FC-CLIP~\cite{yu2023fcclip}). DouC steadies frozen CLIP by fusing
two frozen branches \chg{and never trains}; we \chg{train a readout, then give its
frozen evidence back}, undoing the collapse \chg{the} training made. \chgp{Both
families are} judged by a single mIoU, and which
classes can be trusted is never measured.

\paragraph{\chg{The statistic is old; the use is new.}}
\looseness=-2
The max-cosine of a class name to the training vocabulary is a known statistic put
to other uses: OVSNet~\cite{ovsnet2025} builds hard benchmarks from it,
OVSeg~\cite{liang2023openv} splits seen from unseen classes by hand,
FG-OVD~\cite{bianchi2024fgovd} varies the query names in detection, and
MESS~\cite{blumenstiel2023mess} even reports the training-vocabulary variant's
positive correlation with per-class IoU, as a benchmark characterization,
never as a predictor. DAC~\cite{wang2024dac} calibrates fine-tuned CLIP
classifiers by the same proximity, and at $K{=}1$ its score is
order-equivalent to ours. Name engineering has filled the deployment
side~\cite{pratt2023cupl,menon2023dclip,roth2023waffleclip,yang2023labo};
on the training side, the vocabulary has never been an experimental
axis~\cite{wu2024openvocab}: its membership has never been intervened on
with the images held fixed. \chg{Here the coordinate does three new things: it}
predict\chg{s} per-class collapse from text before any image is seen, \chg{tests} the list
as a cause, and \chg{prices} a certificate. Risk control certifies samples and
pixels~\cite{angelopoulos2022crc,bates2021rcps}; ensemble uncertainty filters \chg{individual segmentations}
at population scale~\cite{norajitra2027ensemble}; we
certify classes, over an entire deployed vocabulary.
The readout--representation gap we find in dense prediction is surfacing
concurrently in other modalities: negation, spatial relations, VQA
utilization~\cite{lu2026negation,li2026readoutblind,veikherman2026encoded}.
A 3D-fusion cousin keeps what argmax drops by retaining the top-$k$
ladder~\cite{gah2026beyond}.

\section{\chg{One readout, one text similarity, one audit protocol}}
\label{sec:method}

\chgp{Which classes can be trusted is measurable before deployment.}

\begin{figure*}[t]
  \centering
  \includegraphics[width=0.97\textwidth]{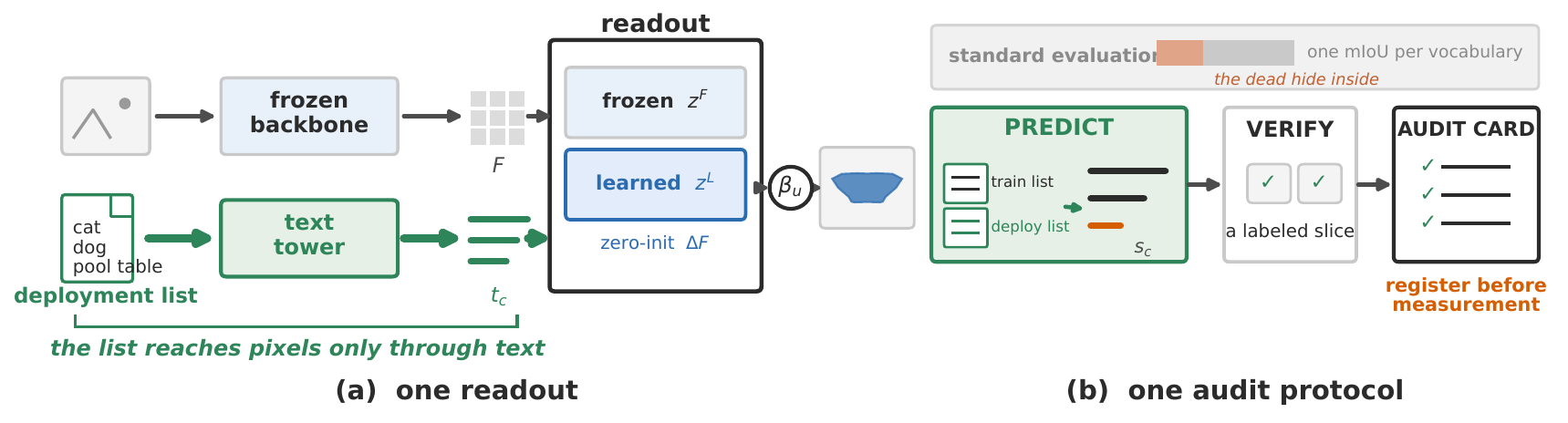}
  \caption{\textbf{The audit setup.} (a) One readout: the class list
  reaches pixels only through text embeddings. (b) One audit protocol: rank the
  names from the text file, verify with a labeled slice of the list, register the card before
  measurement.}
  \label{fig:method}
\end{figure*}

\begin{figure}[t]
  \centering
  \includegraphics[width=0.90\linewidth]{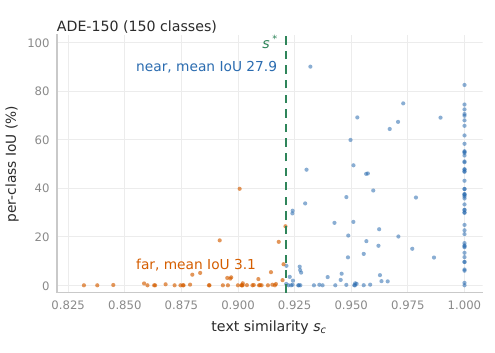}
  \caption{\textbf{Text alone separates kept from collapsed.}}
  \label{fig:scaxis_main}
\end{figure}

\paragraph{One readout, trained once.}
\looseness=-2
Given image $x$ and target vocabulary $\mathcal{V}$, a frozen
spatially-aware encoder (TIPSv2~\cite{tips2026v2}) gives dense features $F$
and a text encoder gives unit class vectors $t_c$. \chgp{The readout is built
to isolate what training changes (Fig.~\ref{fig:method}a): a \emph{frozen
branch} keeps the pretrained scores fixed, and a \emph{learning branch} adds a
small correction through a zero-initialized residual. Both branches score
pixel--class pairs by a scaled cosine to the name's text vector. Training
starts from the frozen readout, and every deviation from it is training's own
effect.} One training run on COCO-Stuff-171\chgp{~\cite{lin2014coco,caesar2018coco}} serves every
deployment\chg{. The} readout is evaluated on deployment vocabularies drawn from
four source datasets (ADE20K\chgp{~\cite{zhou2017ade20k}}, Pascal
Context\chgp{~\cite{mottaghi2014role}}, PASCAL
VOC\chgp{~\cite{everingham2010pascal}}, Cityscapes\chgp{~\cite{cordts2016cityscapes}}).

\paragraph{\chg{One text similarity, one gate.}}
\looseness=-2
Which classes does a trained readout keep? The answer must come before any
image is seen, so we look where readout and vocabulary meet\chg{, the} text
embedding. A target class's \emph{text similarity} to the training vocabulary is
the cosine to its \emph{nearest training name},
\begin{equation}
s_c=\max_{j}\;\cos\big(t_c,\,t_j^{\mathrm{train}}\big)\chg{.}
\end{equation}
\chg{A} class with a near name in the list inherits that name's supervision; a
class with none stands alone (Fig.~\ref{fig:scaxis_main}). \chg{The ranking has a break-even point, $s^*$; below it a class is far, above it near.} \chgp{The learned branch's correction only reaches the seen names. The} further a vocabulary sits from the training list, the more the
\chgp{readout leans on the frozen branch}, so the two branches'
scores fuse as $z_{u,c}=z^L_{u,c}+\beta_u z^F_{u,c}$\chgp{. The best $\beta_u$
per vocabulary, the oracle, drifts with that distance, and finding it would
need target labels. The gate instead reads $\beta_u$ off the learned
branch's own scores: a}
pixel unsure of every training name leans on the frozen branch. \chgp{Set once
on source held-out classes, the gate matches}
the per-vocabulary oracle within $0.3$ mIoU\chg{,} \chgp{without any target labels}.

\paragraph{\chgp{KeepCert, the audit protocol,} predicts from text and verifies on a labeled slice.}
\looseness=-2
Given a frozen backbone, a readout trained on a source seen vocabulary
$\mathcal{V}_{\mathrm{train}}$, and a \emph{target} vocabulary $\mathcal{V}$,
the audit returns a subset $A\subseteq\mathcal{V}$ and a certificate. At
least a $1{-}\alpha$ share of the accepted classes clear a per-class IoU bar
with confidence $1{-}\delta$, and the random unit is the \emph{class}.
\chgp{KeepCert} delivers the \emph{relative} form of this promise\chgp{: a certified lift
over the vocabulary's base rate. The absolute form is priced in the
experiments}. The protocol is the predict-then-audit split of audited conformal
prediction~\cite{zhou2026audited}, instantiated per class
(Fig.~\ref{fig:method}b). Predict reads only the text\chg{, ranking} every
target class by $s_c$ \chg{and proposing} the nested candidate sets
$A_t=\{c:s_c\ge t\}$. \chgp{Setting
$t=s^*$ from the list alone misses the nominal level on held-out classes.}
Verify spends labels\chg{, labeling} $m$ calibration
classes\chg{, choosing} $t$ on them alone\chg{, and certifying} the rest out of sample,
with $m$ the certificate's explicit price~\cite{kossen2021active}.
Text supplies the \chg{ranking}, labels buy the guarantee.

\section{Experiments}
\label{sec:exp}

\looseness=-2 Throughout, a class counts as \emph{collapsed} when its
adapter-only IoU falls below \chg{$1\%$}\chgp{. On ADE-150, half of the collapsed
classes win exactly zero pixels}. \chg{Training trades far classes for near ones (Tab.~\ref{tab:main}).} The average can look healthier as coverage dies\chgp{: on a second backbone (TIPS-B/14), the ADE-150 average climbs from $15.0$ to $20.8$ while its dead-class count rises from $12$ to $39$.} On ADE-150, $28$ of the $51$ dead read above $10$ IoU
before training\chg{. They were alive before training killed them}. Full derivations,
per-cell tables, protocols and controls \chg{are} in the supplementary material.
\begin{table*}[t]
  \centering\footnotesize
  \caption{\textbf{Training trades far classes for near ones, and the divide is field-wide.} Cross-vocabulary mIoU (\%). Frozen row:
  \emph{value} readout, trained row: \emph{patch}. Third
  row: the same run stopped early (epoch $1$). Fourth row: classes below $1\%$ IoU
  before and after training.}
  \label{tab:main}
  \setlength{\tabcolsep}{6pt}
  \renewcommand{\arraystretch}{0.92}
  \setlength{\aboverulesep}{0.3ex}
  \setlength{\belowrulesep}{0.5ex}
  \begin{tabular}{lcccccc}
    \toprule
    & \multicolumn{2}{c}{seen / near} & \multicolumn{4}{c}{far}\\
    \cmidrule(lr){2-3}\cmidrule(lr){4-7}
    method (backbone, train) & PAS-20 & PC-59 & ADE-150 & PC-459 & A-847 & OpenBench\newc{~\cite{ovsnet2025}}\\
    \midrule
    frozen readout (TIPS-L/14) & 82.5 & 36.1 & 23.7 & 8.6 & 9.9 & 28.2\\
    trained readout (TIPS-L/14, COCO-171) & 95.8 & 54.5 & 20.5 & 9.1 & 4.5 & 9.7\\
    \rowcolor{gray!15} \quad \emph{stopped early} & 90.0 & 47.9 & 28.3 & 13.2 & 11.2 & 23.0\\
    \rowcolor{gray!15} \quad \emph{classes scoring below 1\% IoU (before $\to$ after training)} & $0{\to}0$ & $0{\to}1$ & $12{\to}51$ & $122{\to}223$ & $394{\to}679$ & $14{\to}70$\\
    \midrule
    PEARL (CLIP-B, free)~\cite{pearl2026} & 86.9 & 38.5 & 19.4 & 7.7 & 7.1 & 26.0\\
    CorrCLIP (MetaCLIP-L, free)~\cite{corrclip2025} & 91.5 & 50.8 & 30.7 & 13.8 & 13.4 & 34.7\\
    CAT-Seg (CLIP-L, ft)\newc{~\cite{cho2024catseg}} & 97.0 & 63.3 & 37.8 & 23.8 & 15.9 & 35.5\\
    SED (ConvNeXt-L, ft)\newc{~\cite{xie2024sed}} & 96.1 & 60.9 & 35.2 & 22.1 & 13.7 & 33.5\\
    FC-CLIP (ConvNeXt-L, ft)\newc{~\cite{yu2023fcclip}} & 94.7 & 48.2 & 28.4 & 12.4 & 13.2 & 29.4\\
    \bottomrule
  \end{tabular}
\end{table*}

\subsection{Text similarity \chg{alone} ranks collapse risk before any image is seen}
\label{sec:exp_predict}

\begin{table}[t]
  \centering\footnotesize
  \caption{\textbf{Text similarity ranks collapse on nine families, and the
  ranking transfers to two more measured only after the predictions were
  written down.} Each cell spans the row's backbones and the far
  vocabularies (ADE-150, PC-459, A-847; two for Jina-CLIP-v2). $\star$:
  blind families; $^{\dagger}$\,held-out checkpoint of a covered recipe.}
  \label{tab:support}
  \setlength{\tabcolsep}{3pt}
  \renewcommand{\arraystretch}{0.92}
  \setlength{\aboverulesep}{0.3ex}
  \setlength{\belowrulesep}{0.5ex}
  \resizebox{\columnwidth}{!}{%
  \begin{tabular}{lccc}
    \toprule
    & & \multicolumn{2}{c}{collapse rate}\\
    \cmidrule(lr){3-4}
    recipe (backbones) & AUROC & far & near\\
    \midrule
    self-distillation (TIPS-B/L/g/14) & \newc{$0.69$}--$0.88$ & $0.56$--$0.92$ & $0.18$--\newc{$0.65$}\\
    softmax contrastive (CLIP-L, OpenCLIP-B) & $0.86$--$0.91$ & $0.74$--$0.99$ & $0.34$--\newc{$0.80$}\\
    convolutional (ConvNeXt-L) & $0.73$--$0.84$ & $0.49$--$0.87$ & $0.09$--$0.47$\\
    sigmoid (SigLIP-SO400M, SigLIP2$^{\dagger}$) & $0.86$--$0.96$ & $0.90$--$0.99$ & $0.38$--$0.84$\\
    masked reconstruction (EVA02-L) & $0.84$--$0.89$ & $0.92$--$0.99$ & $0.43$--$0.86$\\
    spatial (PE-Core-L) & $0.83$--$0.88$ & $0.93$--$0.99$ & $0.62$--$0.91$\\
    data-curated (MetaCLIP-B) & $0.85$--$0.90$ & $0.78$--$0.99$ & $0.42$--$0.83$\\
    masked multimodal (FLAVA) & $0.83$--\newc{$0.86$} & $0.90$--$1.00$ & $0.48$--$0.88$\\
    \midrule
    AIMv2-L/14$^{\star}$ & $0.81$--$0.87$ & $0.79$--$0.98$ & $0.31$--$0.82$\\
    Jina-CLIP-v2$^{\star}$ & $0.87$--$0.88$ & $0.58$--$0.89$ & $0.05$--$0.21$\\
    \bottomrule
  \end{tabular}}
\end{table}

\begin{figure}[t]
  \centering
  \includegraphics[width=0.88\linewidth]{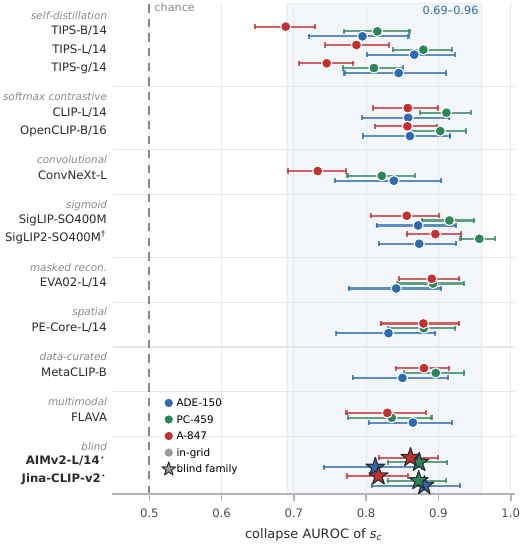}
  \caption{\textbf{Every cell of the bench, ranked from text alone.}
  Whiskers: 95\% bootstrap CIs.}
  \label{fig:forest_main}
\end{figure}

\paragraph{\chg{The text similarity outranks every cheaper coordinate.}}
\looseness=-2 Across nine families, twelve backbones and eight recipes,
the text similarity\chg{, computed from the two lists alone,} ranks collapse at AUROC $0.69$--$0.96$
(Tab.~\ref{tab:support}, Fig.~\ref{fig:forest_main})\chg{. }\chgp{Below the
break-even point, collapse is the norm; on the
near side IoU grows with the similarity.} \chg{Pretraining}
concept frequency~\cite{udandarao2024zero} trails by $+0.21$ AUROC on whether a
class is ever emitted ($11/11$ held-out folds), string overlap \chg{falls}
further behind (Tab.~\ref{tab:rivals})\chg{. The} strongest visual alternative \chg{needs} labels that
do not exist at audit time, and still loses. \chgp{The ranking is not a
seen-versus-unseen split: within the classes that have no identifiable
training counterpart at all, the coordinate still ranks the collapse (mean
AUROC $0.71$).} \chg{Selection asks more than
ranking~\cite{norajitra2027ensemble}. On all $42$ cells, the kept set is
strictly cleaner than the full vocabulary at every coverage below $100\%$.} \chg{Deployed with the kept set, no class dies silently.}

\begin{table}[t]
  \centering\footnotesize
  \caption{\textbf{No rival coordinate beats the text similarity.} Collapse
  AUROC (L/14). The centroid row is paired with $s_c$ on its
  centroid-qualified subset ($s_c$ there: $0.87$/$0.91$). On the full
  30-cell grid nothing beats max-cos significantly.}
  \label{tab:rivals}
  \setlength{\tabcolsep}{3pt}
  \renewcommand{\arraystretch}{0.92}
  \setlength{\aboverulesep}{0.3ex}
  \setlength{\belowrulesep}{0.5ex}
  \begin{tabular}{@{}lccl@{}}
    \toprule
    coordinate & ADE-150 & PC-459 & needs beyond the two lists\\
    \midrule
    \textbf{$s_c$ (max-cos; ours)} & \textbf{0.87} & \textbf{0.88} & ---\\
    visual-distance centroid & $0.79$ & $0.76$ & GT masks + labels\\
    concept frequency & $0.49$ & $0.69$ & ---\\
    string overlap (Jaccard) & $0.55$ & $0.43$ & ---\\
    name length & $0.38$ & $0.34$ & ---\\
    \bottomrule
  \end{tabular}
\end{table}

\paragraph{\chgp{It transfers across four boundaries it was never tuned for.}}
\looseness=-2 \chg{Published pipelines we never trained keep it (SED~\cite{xie2024sed}
$0.68$--$0.70$, FC-CLIP $0.70$--$0.74$ AUROC). Deployment scale keeps it
(LVIS-1203, five towers, $0.72$--$0.88$~\cite{gupta2019lvis}). A second
task keeps it, open-vocabulary detection~\cite{minderer2022owlvit}.
Translated into Chinese, Spanish, French or German, the same deployment
list keeps it ($0.82$--$0.88$). The ordering is the lists' own, and
transfers with them.}

\paragraph{Text alone also counts the \chg{collapse}; \chg{three frontier LLMs fall} short.}
\looseness=-2 Across $64$ held-out cells, the collapsed share tracks a text-computed far mass
at Spearman $0.905$ with slope $1.01$, nearly the rate itself, while
shuffled lists give $0.017$. Three frontier LLMs shown
only the two name lists \chg{fall short of a free cosine}, and spelling the mechanism out to them makes them worse. \chg{The list alone carries it}.

\paragraph{A predictor can be wrong, be calibrated, and then transfer; a report cannot.}
\looseness=-2 The prediction was registered before measurement, and held
blind on a new checkpoint of a covered recipe and on two families measured
only after the predictions were written down. \chg{It} failed twice on LVIS-1203 (within-benchmark superiority does not transfer)\chg{, was corrected, and re-registered}. \chg{It has held since: first} on PC-459, then
$13/13$ blind at both ends on a deterministic split of A-847, and $13/13$
again from an external registry, written down before the complement
vocabulary was ever measured. \chg{Our prediction has been wrong, been calibrated, and transferred.
Which classes collapse, on whose pipeline, in which language, and how
many. The list answers all four.}

\subsection{A class's gain flips by 21 IoU when its counterpart leaves training}
\label{sec:exp_law}

\begin{figure}[t]
  \centering
  \includegraphics[width=0.90\linewidth]{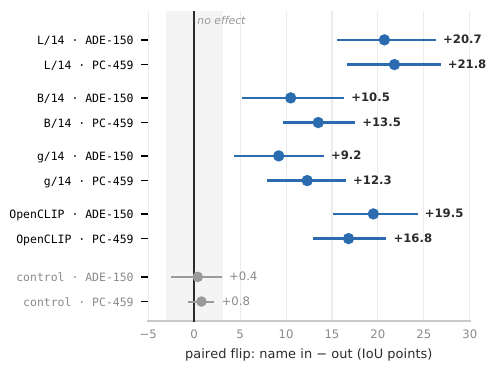}
  \caption{\textbf{One name decides the flip.} Identical images,
  architecture, and recipe; only the counterpart's membership in the
  training vocabulary moves.}
  \label{fig:causal_flip}
\end{figure}

\paragraph{Same images, same recipe\chg{. Only} the name's membership moves.}
\looseness=-2 We run the first intervention on a training vocabulary alone
(Fig.~\ref{fig:causal_flip}). COCO-171 is split into two disjoint halves, one
readout trained per half on the same images with the other half's pixels
ignored; for a target class whose COCO counterpart is identifiable
($\cos\ge0.97$), the paired difference between the two readouts isolates
that name's membership in the list, so
the VDM is a consequence of the \chg{training list}.
The flip is large and asymmetric\chg{. Membership} holds the
class at the frozen readout's level ($+2.6$)\chg{; absence} destroys it ($-18.2$)\chg{. The name's power falls on the absent.} A
control arm swapping in unrelated names, same size, different composition, leaves the gain at zero.

\chg{The name and its pixels act as a pair, and each fails alone. Far
names appended to the list without pixels never learn to win, and every
one collapses (Tab.~\ref{tab:flip}). With every pixel still supervised,
swapping the training name for a far-embedded substitute collapses the
class\chgp{, and all twenty substitutes inherit the pixels}.
Membership also rescues\chgp{: re-}adding the predicted classes' counterparts to
the list halves the harm exactly on the named half and leaves the rest
untouched. \chgp{The same effect holds at single-name granularity: removing one
name from the training list kills that class, a paired drop of $39$ to $86$
IoU points across the four we tested. The lever reaches past the removed
class: removing the single dominant name \emph{sky} from training lifts the
adapter-only mIoU on ADE-150 from $20.5$ to $46.9$, because that class had
been eating the others' pixels.} An edit to a text file before training decides which classes
the readout keeps.}

\begin{table}[t]
  \centering\footnotesize
  \caption{\textbf{A second list writes the same flip.} Paired gain
  difference, counterpart in $-$ out of the training half (IoU points,
  L/14; images and recipe fixed). Every 95\% CI excludes zero.}
  \label{tab:flip}
  \setlength{\tabcolsep}{5pt}
  \renewcommand{\arraystretch}{0.92}
  \setlength{\aboverulesep}{0.3ex}
  \setlength{\belowrulesep}{0.5ex}
  \begin{tabular*}{\columnwidth}{@{\extracolsep{\fill}}lr@{}}
    \toprule
    \multicolumn{2}{@{}l}{\emph{ADE-150 halved --- registered before training}}\\
    \quad$\to$ PC-459 & $+11.6$\\
    \quad$\to$ A-847 & $+21.3$\\
    \quad$\to$ COCO-171 & $+10.6$\\
    \midrule
    \multicolumn{2}{@{}l}{\emph{far names appended, no supervision: all $43$ collapse ($\Delta$ $-2.5$)}}\\
    \bottomrule
  \end{tabular*}
\end{table}

\section{The Vocabulary-Distance Mechanism}
\label{sec:analysis}

\chg{Training loss falls} the whole time the far band
dies. \chgp{The VDM makes the two curves
one process, in four pieces, each carrying a way to kill it.}
$\mathcal S=\mathrm{span}\{t^{\mathrm{train}}_j\}$ is the training-text span.

\subsection{\chg{Why text alone \chgp{predicts} which classes collapse}\chgp{. Training writes only in} \texorpdfstring{$\mathcal S$}{S}}
\looseness=-2
\emph{Kill it} by \chg{finding one class outside the span that training improves}.
\chg{Under softmax cross-entropy over the source seen classes, the update
to a pixel's feature is a residual-weighted combination of the training
text vectors, $\Delta F_u=\sum_j (p_{u,j}-y_{u,j})\,t_j^{\mathrm{train}}$,
so it lies in their span\chgp{ for a dot-product readout. The scaled cosine adds a
projection orthogonal to the current feature, and the update's in-span energy
share stays above $99.8\%$}. A deployment-only class is never the label, so
its positive-term gradient is identically zero and no training-loss curve
can see it collapse.}

\begin{figure*}[t]
  \centering
  \includegraphics[width=0.97\textwidth]{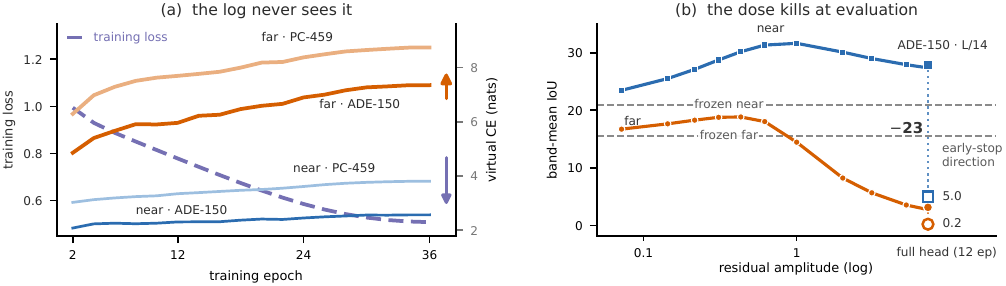}
  \caption{\textbf{\chgp{One dose axis, two views.}} \chgp{(a) Training loss
  falls while the far band's own loss, never optimized, rises. (b) At
  evaluation the same axis reads as the residual's amplitude. The far band
  dies along it, and at full dose the early direction alone flips both
  bands (open markers).}}
  \label{fig:dose}
\end{figure*}

\noindent \chg{Only the part of a deployment class's name that overlaps the
training span drags the class.} On a frozen backbone the readout is a linear classifier on fixed
features, and such classifiers keep growing in norm as $\Theta(\log t)$ long after the
direction has
converged~\cite{soudry2018implicit,lyu2020margin,ji2020directional}. \chgp{The scalar loss therefore
falls while the far band's virtual cross-entropy (the true class's probability on its own pixels, which training never optimizes) rises} (Fig.~\ref{fig:dose}a).
\chg{On two of six benchmarks, A-847 and a
rebuilt stress set}\chgp{~\cite{ovsnet2025}}\chg{,} the frozen
readout's far band \emph{beats} its near band by $+1.5$ and $+9.9$ IoU\chg{, while} every trained readout orders near over far on all six\chg{. Training}
writes the distance ordering.
\chgp{Holding} images and pixel
supervision fixed, \chg{we} swap only which \emph{name} each of twenty training-class
pairs is supervised under. At evaluation the pixels follow the name the
class trained under (capture $73.2\%$), not the semantically correct one
($0.3\%$). A class's fate is its position relative to the list.

\begin{figure}[t]
  \centering
  \includegraphics[width=0.95\linewidth]{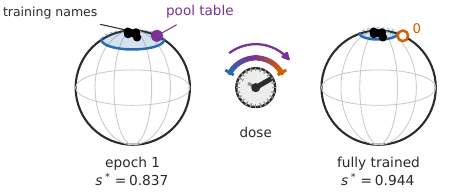}
  \caption{\textbf{Training closes the cap the list keeps alive.}}
  \label{fig:sphere}
\end{figure}

\paragraph{The forecast. The margin model turns position into a forecast, and the break-even is the list's.}
\looseness=-2
Each training name keeps alive a spherical cap on the name-embedding sphere
(Fig.~\ref{fig:sphere}), whose half-angle is $\arccos s^*$\chg{. Inside} a cap, a
class lives; outside all caps, it dies. Formally, if training amplifies the
in-span directions at rate $\kappa$ and attenuates the rest at rate
$\lambda$, the per-class \emph{margin} gain is affine with a single zero
crossing,
\begin{equation}
\mathrm{gain}_c\;\approx\;a\,(s_c-s^*),\qquad s^*=\lambda/\kappa\chgp{,}
\label{eq:law}
\end{equation}
\emph{\chg{so }a class is predicted to collapse when its text similarity to
the training vocabulary falls below the break-even point}.
The whole geometry then reduces to one question\chgp{:} where the break-even sits. \chg{The} list answers it \chg{from its own structure}\chg{. How} tightly \chg{its} own
names cluster predicts $s^*$ to within $0.013$ on average across nine
families (worst family $0.04$), and \chg{the prediction, written down before either was measured, holds on} the two further \chg{families} (AIMv2, Jina). Text alone forecasts both
which classes die and where the break-even sits.

\chgp{The break-even cannot be moved without touching the list's geometry or
the dose.} \chg{The choice of names cannot move it, nor the optimizer's path,
nor the frame of the space;} only the dose moves it, and, within the list, only its size, with diminishing returns.
One object underlies all of this\chg{, a} spherical cap the list keeps alive,
whose radius the list itself sets. The cap is a predictor, not only a description. We wrote down the break-even of $38$ never-measured heads
before evaluating any of them, and all $38$ landed within ${\pm}0.02$; at
negative doses the relation inverts\chg{, and} on the far-dominated vocabulary the
gain ordering flips sign and the \chgp{collapse} list empties. \chg{The cap pins the
band, not the point.}

\subsection{\chg{Why one knob repairs the collapse}. The damage concentrates on one amplitude axis}
\looseness=-2
\emph{Kill it} with one effective repair that leaves amplitude free.
What training keeps growing, once the direction has converged, is the margin's amplitude, so the dose is the residual amplitude\chg{.} \emph{\chg{How} much}
a head collapses is set by $\|\Delta F\|$ (\chg{monotone within each of
three backbones, $\rho=-1.00/-1.00/-0.90$,} over fourteen heads); the
repair/damage divide sits at unit relative
amplitude $\mathrm{RMS}(\Delta F)/\mathrm{RMS}(F){\approx}1$\chg{, the} scale where
the training equilibrium balances weight growth against
decay~\cite{kosson2024rotational}. \newc{When the residual is held near the
frozen scale during training, its amplitude crests at $1.5$ and settles at
$1.02$, on the divide itself. At this dose the far band dips as the
amplitude crests and returns as it settles (rank correlation $-0.78$ along
the run), while at full dose the damage remains after the amplitude
recedes. Two of the three deployment vocabularies end the held-down run
within $7\%$ of their first-epoch far level; the third keeps a slow leak.} The dose also moves the break-even, not
only the damage\chg{. At} fixed recipe it climbs with the residual dose and saturates. The cap closes as training
proceeds, and it closes in text order, far classes first. Two epochs of
training already suffice\chg{. The} per-class virtual cross-entropy at epoch $2$
ranks the eventual dead at AUROC $0.930$ (ADE-150) and $0.926$ (PC-459).
\chg{The} largest drop lands in the
first interval, and the decay is monotone from there. Collapse is immediate
specialization. Early
stopping therefore pulls the cap back\chg{. The} first two \chg{pieces} couple exactly
where the damage begins.

\paragraph{The rival. \chg{Direction modulates, amplitude decides.}}
\looseness=-2
\chgp{At} one and the same amplitude, the fully trained head's own, swapping the residual's
direction from the final checkpoint's to the early checkpoint's lands both
bands at the opposite end (Fig.~\ref{fig:dose}b). Direction alone costs $23$ IoU, more than
the repair recovers.
\chg{The directions that spare far classes
live} only on the early, low-amplitude checkpoints. Reaching them is what
turning the amplitude down \emph{does}.
\textbf{Amplitude is the dose axis\chgp{; it decides which
directions are reachable}; direction only modulates
the damage at a fixed dose.} Necessity, not just correlation\chg{.
Zeroing} the update's out-of-span component leaves the collapse intact,
while every effective repair constrains amplitude. The vehicle is \chg{in-span amplitude growth}.
The same knob writes the harm and its predictability at once\chg{. At} zero trained
amplitude the \chgp{collapse} list is anti-ordered (collapse AUROC below
chance), and predictability grows monotonically with the dose.
Amplitude decides how much damage; it does
not decide where the lost pixels go.

\subsection{\chg{Where the lost pixels go}. Attractor basins set by text geometry and class mass}
\label{sec:link3}
\looseness=-2
\emph{Kill it} by predicting the sinks from the images instead, or by
excising the sinks and returning the pixels.
\chgp{Far pixels funnel into the} attractor classes of the training vocabulary, the sinks of the text manifold (cf.\ neural
collapse~\cite{papyan2020neuralcollapse}; \chg{the hubness cousin in
zero-shot
classification~\cite{radovanovic2010hubs,dinu2015improving,sato2026hubness}
offers no account of per-class collapse in a deployed vocabulary}). \chgp{Attractor
load tracks two inseparable factors, text centrality and the pixel mass
itself.} Text geometry nominates the candidates, class mass picks the winners. \chg{Excising} the top attractor names, the highest-mass names,
or random names at \emph{training} time \chgp{leaves the collapse unchanged}
(all CIs through zero). The sink's identity is replaceable,
only the span must have one. The same names recur as sinks across towers,
and
the division of labor is clean\chg{. Text} distance decides who becomes
vulnerable, visual proximity and class mass decide who eats the pixels;
\chgp{the sinks are steerable from}
the list alone.
The image side offers no competing account\chg{. Corrupting} the images outright
moves the near$-$far gap by ${\le}1.8$ points against an $11.7$--$24.8$
vocabulary-side span, and masking the top-12 measured attractors returns
almost none of the collapsed pixels. \chgp{The basin re-forms around the next
names.}

\subsection{\chg{What collapse is}\newc{. Every vote lost, the features survive}}
\looseness=-2
\emph{Kill it} with a probe that reads nothing, or a decoy that steals
nothing.
Each pixel goes to the highest-scoring name\chgp{: the} deployment vocabulary's
classes compete\chgp{, and} a collapsed class is one that never
wins. \chg{An} oracle linear readout on the
same frozen patches reaches
$30.4$ IoU over the far band\chg{, nearly double the frozen readout's
level. On the collapsed classes it reaches} $28.6$ \chg{where} the
trained readout scores
$0.11$ (shuffle control ${\approx}0$); a wider probe battery
agrees~\cite{alain2016probes,davari2022probing}, supervised or training-free. The features carry what the task needs; the trained readout takes almost none of it. Fed the collapsed
classes' own pixels, the deployed readout
returns the true class for almost none of them,
and the name that wins instead is a near-vocabulary neighbor in
$93\%$/$96\%$ of cases (a palm's pixels go to \emph{tree}, a pool
table's to \emph{table}). The pixels are stolen by near names.
\chgp{The theft is silent. Below} $s^*$ the readout's top-1 confidence stays
$0.52$ while the true class's own score sits at $0.038$\chg{. The} confidence is read off the nearest \emph{seen} name,
\begin{equation}
\max_{c'}\,p^L_{u,c'}\;\approx\;\phi(s_c),\qquad \phi\ \text{monotone
increasing}
\label{eq:conf}
\end{equation}, a map of how seen-like the pixel is, not of whether it is right. The
same silence replicates on the rebuilt stress set, \chgp{where the far band's
confidence sits at $0.45$ against the frozen readout's $0.67$ while its
win rate falls from $0.44$ to $0.04$}.
Collapse is the mirror
image of neural collapse~\cite{papyan2020neuralcollapse}: there, training
collapses the features onto class means; here the features stay readable and
the assignment collapses onto the training vocabulary's nearest names.
\paragraph{The attack. The competition can be stolen.}
\looseness=-2
\chgp{The competition is also a regulator: a}
BCE readout with no softmax competition collapses
\emph{every} far class and halves the near ones (two families). What runs on names
can be stolen from the text file alone\chg{. One} near-name decoy appended to
the deployment vocabulary, no training, no weights, no image or model output queried (an image-level cousin exists in open-set classification~\cite{ren2023openness}), drains its victim class at
inference, the stolen IoU growing monotonically with the decoy's text
proximity ($+8.49$; controls null). \chgp{Attackers who never see the
coordinate rediscover the near-name region on their own. An exhaustive sweep
over the 1,459 real names, ranked by the model's response alone, lands its
successful attacks at median cosine $0.95$ and drains $13.1$ IoU; a
gradient-guided search over three-word strings~\cite{ebrahimi2018hotflip,shin2020autoprompt,zou2023universal} drains $24.2$,
$93\%$ of what a free continuous vector can steal. The region does not bound
crafted blends: $12$ of $150$ searches steal from outside it (max $31.7$), a
second attack surface the coordinate does not cover.} The same one line
drains CAT-Seg, SED and FC-CLIP, and even a promptable concept decoder
with no native competition (SAM 3~\cite{carion2025sam3}), once victim and
decoy are made to compete on the same weights; even the frozen readout is drained. The attack surface is the
shared-vocabulary competition
itself, and training only sharpens it. Renames steer the same competition
in both directions\chg{. A} collapsed class revives without touching an image
(\emph{flag}${\to}$\emph{banner}\chg{,} $0{\to}28.5$ IoU), and a healthy class
drains under a farther synonym (\emph{curtain}${\to}$\emph{drape}\chg{,}
$54.9{\to}21.6$). At scale ($32$ collapsed classes, three rename rules),
crossing the break-even is neither sufficient nor necessary (\emph{tank}${\to}$\emph{storage tank}
revives while its similarity drops\chgp{)}\chg{. What} revives a class is escaping its
local rival, and the coordinate's blind spot is exactly where that
competition lives.

\section{\chg{One knob undoes the collapse}}
\label{sec:exp_repair}\label{sec:exp_cert}
\begin{figure}[t]
  \centering
  \includegraphics[width=0.90\linewidth]{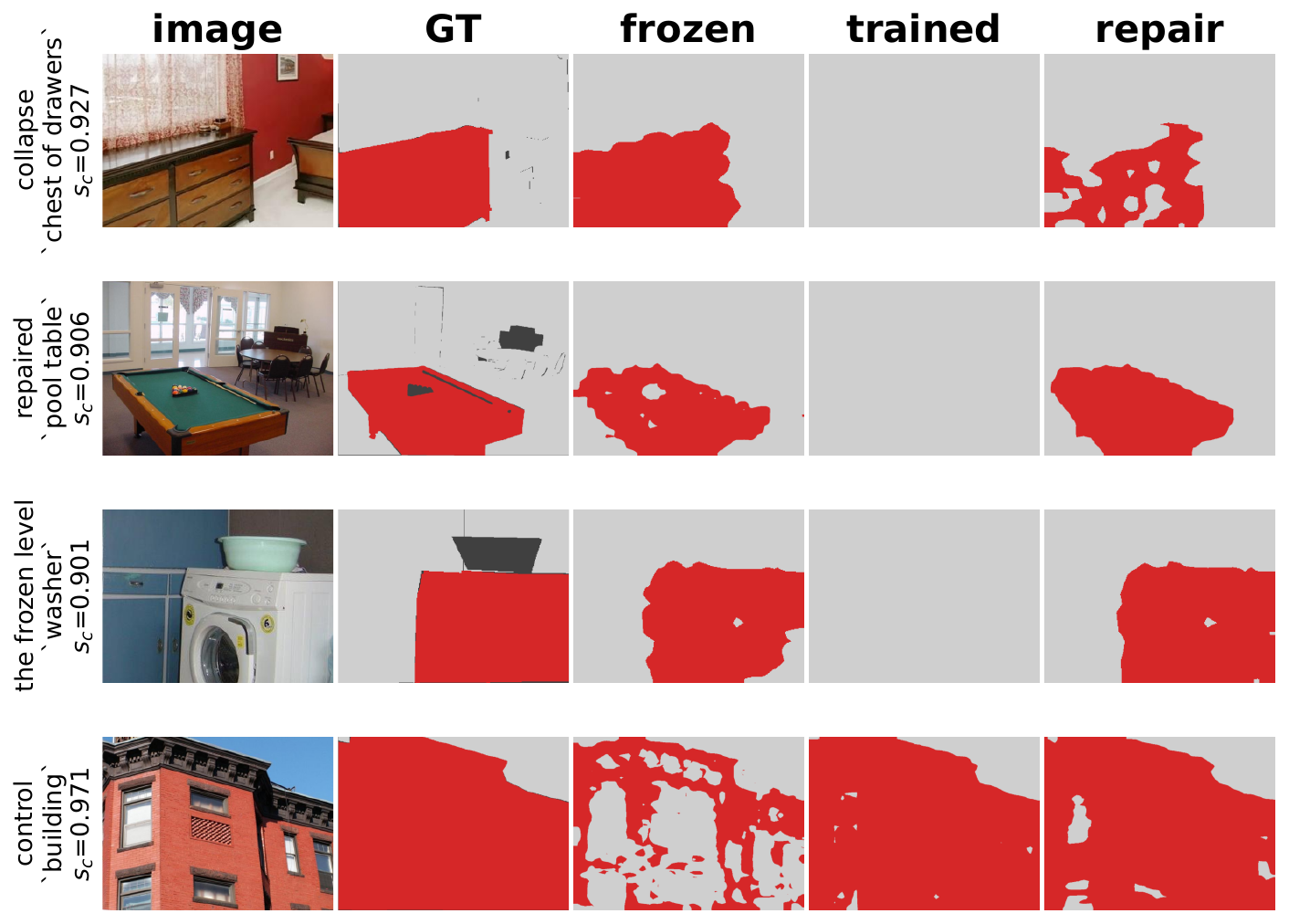}
  \caption{\textbf{Collapse and repair, up close.} Four ADE-150 cases, drawn
  from the classes the coordinate marks for collapse; the repair column turns
  the residual amplitude down, no retraining.}
  \label{fig:masks}
\end{figure}

\paragraph{Turning the amplitude down brings the lost classes back.}
\looseness=-2 \chg{The mechanism predicts its own repair. What training
turns up, turning back down should undo}~\cite{buna2026compute}\chg{. It
does. Stopping training at epoch 1 recovers $+16.2$ far IoU on ADE-150,
and a zero-training variant that projects the residual back to its
epoch-1 amplitude recovers $+15.3$, with no retraining and no labels
(Tab.~\ref{tab:zoo}, Fig.~\ref{fig:masks}). The turn works because the
directions that spare far classes live only on the early, low-amplitude
checkpoints, and turning the amplitude down reaches them.}
The knob returns what the frozen features kept. \chg{Three of the four
repairs bring the far band back to the frozen level or beyond on all
three vocabularies, and the early-stopped readout overtakes the frozen
one on all-vocabulary mIoU on five of the six vocabularies.}

\paragraph{The repair zoo is one knob.}
\looseness=-2 \chg{Four repairs from four literatures (}\emph{early
stopping}, \emph{amplitude projection}, \emph{quadratic penalty},
\emph{residual-amplitude constraint}\chg{) all climb the same axis, and
the} monotone amplitude--collapse relation predicts where each setting
lands. Interpolating back toward the frozen weights is known to
recover average robustness~\cite{wortsman2022wiseft}\chg{. The dose
axis turns that heuristic into a mechanism. It says where to stop, at
unit relative amplitude, and what returns, per-class survival. Off the
axis, nothing moves. Temperature}
scaling~\cite{guo2017calibration} is exactly the fusion-path
coefficient, a monotone rescaling that never reorders a
selection~\cite{rabanser2025selective}\chg{. Logit}
adjustment~\cite{menon2021logit,chao2016gzsl,zhu2023gla} \chg{steers by
class priors, a different axis.} \newc{Adapting at deployment moves the
axis backward. Entropy minimization on the unlabeled
stream~\cite{wang2021tent,daf2026} sharpens the winners, and the far band
sinks further on all three vocabularies while the average falls with
it.} \chg{The knob needs no per-class steering
either. One} global fusion turn beats per-class oracle tuning.

\paragraph{The same knob, installed during training, prevents the
collapse.}
\looseness=-2 \chg{A causal axis works in both directions. Holding the
residual near the frozen scale during training should prevent what
turning it down afterwards repairs\chgp{, and it does. An} L2 brake on} $\Delta
F$\chg{, no labels and no third list, lifts the far band by}
$+8.4$/$+3.1$/$+4.0$ \chg{on the three far vocabularies
(pre-registered). The near side rises with it} ($+2.1$--$+4.5$)\chg{, a
Pareto move the post-hoc knob does not make. A name-bank variant, KL
toward frozen predictions over} $900$ \chg{held-out names, doubles the
far gain to} $+15.4$ \chg{on ADE-150, past the frozen level, and the
same untuned} $\lambda$ \chg{holds on two further backbone families.
Trained on a second list base, the ADE-150 half-vocabularies of the
flip experiment, the name bank gains} $+3.5$--$+5.0$ \chg{on the far
band of both held-out vocabularies.} \chgp{All four pre-registered cells
pass, while steering only the residual's direction changes nothing.
Prevention cannot buy information, and on A-847 even the stronger arm
reaches only the frozen level} ($8.0$ vs $8.7$)\chgp{. Post-hoc knobs
refund to this level, and prevention lands near it. Only supervision adds
what the frozen features do not already carry.}

\begin{table}[t]
  \centering\footnotesize
  \caption{\textbf{One knob undoes the collapse.} Far-band IoU below the
  break-even point (L/14, adapter-only). Top block absolute. Arms below
  are paired $\Delta$ against the full head, blue at or past the
  frozen level. The last row adapts the trained readout on the
  unlabeled deployment stream (test-time adaptation, TTA).}
  \label{tab:zoo}
  \renewcommand{\arraystretch}{0.92}
  \setlength{\aboverulesep}{0.3ex}
  \setlength{\belowrulesep}{0.5ex}
  \begin{tabular*}{\columnwidth}{@{\extracolsep{\fill}}lccc@{}}
    \toprule
    far-band IoU & ADE-150 & PC-459 & A-847\\
    \midrule
    frozen (no training) & $15.59$ & $5.69$ & $8.68$\\
    full head (12 ep) & \textcolor{lossred}{$3.14$} & \textcolor{lossred}{$0.61$} & \textcolor{lossred}{$0.99$}\\
    \midrule
    \multicolumn{4}{@{}l}{\emph{paired $\Delta$far vs full head, four repairs, one axis}}\\
    early stopping (epoch 1) & \textcolor{winblue}{$+16.2$} & \textcolor{winblue}{$+8.1$} & \textcolor{winblue}{$+8.3$}\\
    amplitude projection & \textcolor{winblue}{$+15.3$} & \textcolor{winblue}{$+6.5$} & \textcolor{winblue}{$+8.1$}\\
    residual-amplitude constraint & \textcolor{winblue}{$+15.3$} & \textcolor{winblue}{$+7.1$} & \textcolor{winblue}{$+8.0$}\\
    soft quadratic penalty & $+8.7$ & $+3.3$ & $+4.3$\\
    \multicolumn{4}{@{}l}{\emph{paired $\Delta$far vs full head, the same knob, installed during training}}\\
    residual L2 brake (no labels, no bank) & $+8.4$ & $+3.1$ & $+4.0$\\
    name-bank KL ($900$ held-out names) & \textcolor{winblue}{$+15.4$} & \textcolor{winblue}{$+6.6$} & $+7.3$\\
    \multicolumn{4}{@{}l}{\newc{\emph{off the axis, at deployment}}}\\
    \newc{entropy minimization (TTA)} & \newc{\textcolor{lossred}{$-2.0$}} & \newc{\textcolor{lossred}{$-0.5$}} & \newc{\textcolor{lossred}{$-0.4$}}\\
    \bottomrule
  \end{tabular*}
\end{table}

\paragraph{A lift is certified where it exists and refused where it does not.}
\label{sec:exp_boundary}
\looseness=-2 What the knob cannot buy is the boundary itself\chg{. Spending} the
label budget toward the deployment vocabulary lifts only the near side and
leaves the far side unmoved. The boundary is priced by supervision, not by
budget allocation. \chgp{KeepCert} issues the first certificates of per-class dense risk
on a deployed vocabulary\chg{. On} the far vocabularies the accepted set beats the
vocabulary base rate by $+0.13$ to $+0.28$, a lift certified in $18$--$86\%$
of calibration splits; on the near vocabulary (PC-59) the realized gain is
positive, yet the certificate is withheld in ${\ge}97\%$ of splits, refused, correctly. The absolute form is refused outright, every far cell at $\alpha{=}0.2$ ($0/10$), and the refusal is priced, not incidental\chg{. With} $Q_0,Q_1$ \chg{as} the gate's firing rates on bad
and good classes\chg{, the certificate requires}
\begin{equation}
\begin{split}
R(A) \;\le\; \alpha
\quad&\Longleftrightarrow\quad
Q_1 \;\ge\; c\,Q_0,\\
&\text{with}\quad
c \;=\; \frac{(1-\alpha)\,(1-\pi)}{\alpha\pi}
\end{split}
\label{eq:floor}
\end{equation}
\chg{. At} our base rates $c{\approx}16$/$94$\chg{, the} measured label-free supply
falls ${\sim}6{\times}$ short; a confidence
score at MSP AUROC $0.97$ certifies nothing. Calibration is not certifiability. Labels buy what no label-free score can, at a per-class
price falling with vocabulary size, and the property that makes the attack possible is the one that
makes verification possible\chg{. The} deployment vocabulary arbitrates the
competition.

\section{Conclusion}
\label{sec:conclusion}

\paragraph{The shape of the boundary.}
\looseness=-2 The boundary is measured on both sides\chg{.} \chg{One text similarity and one label budget measure} what a readout keeps, what a knob refunds or prevents, \chg{and} what no label-free score can certify\chg{.} The
scope throughout is the frozen-backbone, trained-readout regime, with a small
labeled calibration set.
An older science asks the same question of a crowded niche\chg{. Coexistence}
thins past a limiting
similarity~\cite{hardin1960competitive,hutchinson1959homage,macarthur1967limiting}, and the break-even falls as the
training vocabulary grows, decelerating, as though the vocabulary metric
admitted only so much separation. A new
vocabulary's risk is readable before training and its certificate priceable
afterwards\chg{. The} boundary is an information budget with a measured
location. The field scores a readout by its average; the average cannot
say which classes die. The training vocabulary itself becomes an object\chg{,} written before
training, read before any image, edited with a text editor.

\section*{LLM Usage Statement}
\looseness=-2\noindent This work is author-driven: the research questions, hypotheses, method
designs, experimental protocols, evaluation criteria, analyses, and conclusions
were conceived, directed, and validated by the authors, who also ran every
experiment. LLM-based assistance (Claude, Anthropic) was limited to language
polishing of author-written text---grammar, phrasing, and style---and
contributed no scientific ideas, code, experiments, results, or
interpretations. The authors reviewed every edit and take full responsibility
for the content.

{
    \small
    \bibliographystyle{ieeenat_fullname}
    \bibliography{main}
}


\end{document}